# HandSight: DeCAF & Improved Fisher Vectors to Classify Clothing Color and Texture with a Finger-Mounted Camera

**Alexander J. Medeiros**

Makeability Lab | HCIL

University of Maryland

College Park, MD, USA

ajmed13@gmail.com

**Lee Stearns**

Makeability Lab | HCIL

University of Maryland

College Park, MD, USA

lstearns86@gmail.com

**Jon E. Froehlich**

Makeability Lab | HCIL

University of Maryland

College Park, MD, USA

jonf@cs.umd.edu

## Abstract

We demonstrate the use of DeCAF and Improved Fisher Vector image features to classify clothing texture. The issue of choosing clothes is a problem for the blind every day. This work attempts to solve the issue with a finger-mounted camera and state-of-the-art classification algorithms. To evaluate our solution, we collected 520 close-up images across 29 pieces of clothing. We contribute (1) the HCTD, an image dataset taken with a NanEyeGS camera, a camera small enough to be mounted on the finger, and (2) evaluations of state-of-the-art recognition algorithms applied to our dataset - achieving an accuracy >95%. Throughout the paper, we will discuss previous work, evaluate the current work, and finally, suggest the project's future direction.

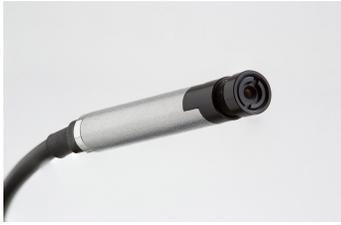

**Figure 1:** A NanEye GS camera - only a few millimeters in diameter.

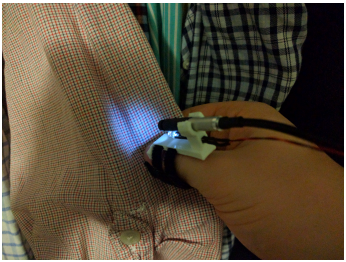

**Figure 2:** A finger-mounted HandSight prototype with an LED. Notice the shadow casted by the user has little impact on the lit area. By blocking the effects of ambient light, we are able to increase accuracy with consistent lighting.

## Introduction

Our research group has been exploring ways to provide visual context of a user's immediate surroundings by augmenting the sense of touch. We call this effort HandSight. This goal is especially important for the visually impaired which accounts for approximately 161 million people globally according to the World Health Organization [7]. The current work adds to HandSight's agenda by facilitating the daily chore of choosing an outfit. Choosing an outfit is a difficult problem in object recognition and user feedback. The latter needs further research to include users who have been blind since birth.

Our work thus far has focused on color and pattern recognition of clothing. We have implemented an image processing pipeline that includes Dense SIFT (DSIFT), DeCAF and Improved Fisher Vectors (IFV) to classify our dataset with >95% accuracy. Moving forward, we will focus on user experience and evaluating different mobile and on-body implementations. We have already worked towards this effort by capturing the HandSight Color Texture Dataset (HCTD) using a NanEyeGS camera (depicted in Figure 1) which is small enough to be mounted on the finger (Figure 2). The HCTD is unique such that it captures close-up images of realistic conditions including varied angle, distance, and tautness of fabric.

Recent research in pattern and texture recognition [1] has produced algorithms that reach >98% accuracy on datasets providing consistent camera settings, while achieving >65% accuracy on datasets such as the DTD [1] and FMD [2]. Our close-up and on-body camera approach generates consistent image characteristics such that our accuracy remains high.

Aside from algorithmic related work, Yang et al. has also explored the problem of choosing an outfit with visual impairment [6]. They produced an algorithm for recognizing clothing color and pattern with ~93% accuracy [6]. In contrast, our classification achieves >95% accuracy with a uniquely mobile solution which includes the finger-mounted camera/LED combo which reduces most ambient lighting. Lastly, our approach classifies more than twice as many textures as Yang's work.

In summary, our contributions include: (1) the HCTD: a unique set of close-up clothing images; and (2) evaluations of state-of-the-art recognition algorithms applied to our dataset - achieving an accuracy >95%. Our findings further validate the use of DeCAF and IFVs as a means of image classification and provide a novel and difficult image dataset to be used with future HandSight development. Further research needs to be done to develop a wireless mobile solution that operates in real-time.

## Background and Related Work

Our texture classification work is largely inspired by the DTD [1] and their success in classifying 47 different textures, including many that accurately categorized clothing patterns such as checkered, striped, floral, etc. Therefore, the 9 categories we chose are derived from the DTD, except for denim and none/solid. The algorithms used for HandSight are adapted from the DTD paper which uses IFVs and DeCAF feature vectors to represent an image. DeCAF vectors have been implemented by the BVLC in the Deep learning framework called Caffe. These were initially used by Krizhevsky et al. (2012) which won the ILSVRC – 2012 [3].

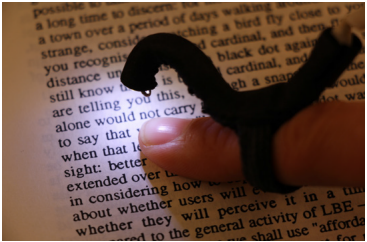

**Figure 3:** One of many HandSight prototypes which facilitates reading for the visually impaired.

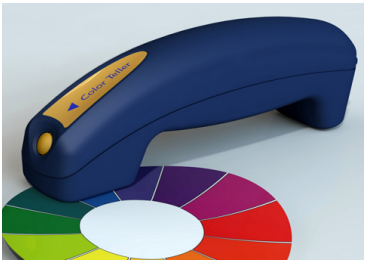

**Figure 4:** Many color grabbers only compute average color rather than the few that are most prevalent in an image.

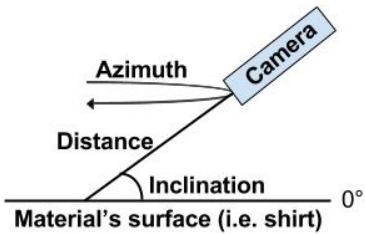

**Figure 5:** Definition of three columns of the HCTD: distance, inclination, and azimuth.

Since that paper, many object recognition tasks have used some variant of the neural network model developed from Krizhevsky's work and found satisfactory results [3]. Furthermore, the DTD evaluated several different feature vector collections including IFV, BOVW, VLAD, LLC, KCB, and DeCAF. They found their highest accuracy with IFV + DeCAF on every dataset they experimented with. Therefore, we went directly to IFV + DeCAF.

In terms of the physical design of HandSight, we used one of our group's prototypes which is evaluated in [4]. In short, HandSight is a finger mountable set of sensors (including a camera) which allows a user to gather visual context through the intuitive sense of "touch." We have already studied HandSight's use for reading text on a page and using gestures as input control for a mobile device as depicted in Figure 3 [4,5]. The current work further explores the ability of finger-mounted sensors to facilitate daily tasks such as choosing clothes.

Matching colors is a key component in choosing outfits. That said, most color identification solutions such as the color grabber in Figure 4 only provide an average color, rather than several specific colors. Our solution can identify an array of colors in a particular image of clothing.

Finally, most similar to the current work, Yang et al. has implemented a system that can classify 4 patterns and 11 clothing colors with ~93% accuracy [6]. They even implemented a speech-to-text controller for sending commands to the system such as "start recognition" and "turn off system." Moreover, they completed a user-study which concluded that blind users desired such a system to support more daily independence [6]. In comparison, the benefits of our approach are evident in low-lighting conditions and in general use. If there is low-lighting for Yang's system, the recognition has limited capability, however, since we've outfitted the camera with an LED, our approach is generally not subject to ambient light. Secondly, Yang's system requires the user to position the clothing to occupy all of the camera's view. To our knowledge, HandSight is the first to use close-up and local features for general classification of clothing color and pattern.

### HandSight Color-Texture Dataset (HCTD)

In order to evaluate our approach we needed to gather a dataset representative of our problem domain (close-up, LED light source, lower resolution camera). We call this the HandSight Color-Texture Dataset (HCTD). The dataset is tabulated as a csv file with the following 11 columns for each image (definitions of several of the columns found in Figure 5):

1. Image ID
2. Label ID
3. Distance between camera and material
4. Angle of camera to material (inclination)
5. Orientation of camera (azimuth)
6. Scale (pixels per cm (PPCM))
7. Lighting (LED configuration)
8. Tension of material (e.g. taut or hanging)
10. Notes
11. Colors

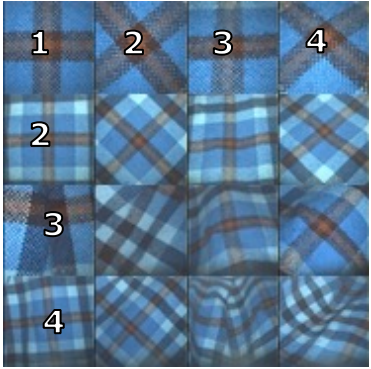

**Figure 6:** The 16 possible configurations considering 2 azimuth angles, 2 distances, 2 inclination angles, and 2 tensions. The 1st and 3rd row are at a distance of 5cm while the others are at 12cm. The 2nd and 4th columns are at an azimuth angle of 45°, while the others are at 90°. The 1st and 2nd rows are taut; the others are hanging on a hook. Finally, the 3rd and 4th columns are at an inclination of 45°; the others are at 90°.

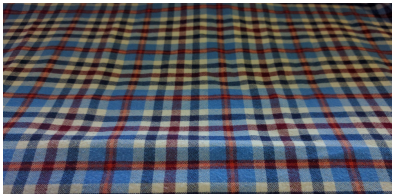

**Figure 7** This is the original texture of figure 6 captured with a Nexus 5X phone camera. The dataset includes an image similar to this for each texture used.

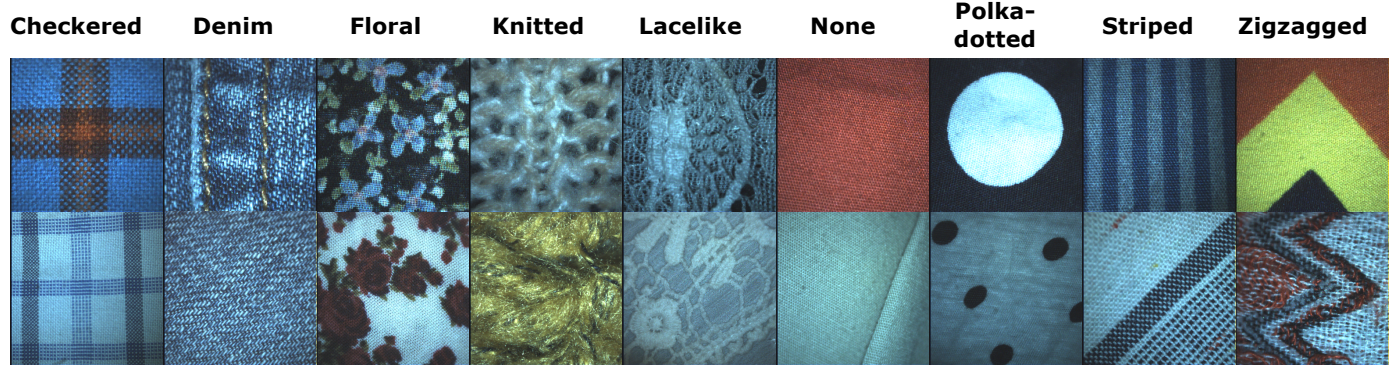

**Figure 8** Depicts each texture in the HCTD. Notice the significant variability between samples of the same texture. Also note the similarities between textures such as denim and none or the second row of none and striped.

(1) The image ID is an integer primary key and (2) the label ID is an integer that maps to a texture type. Distance, inclination, and azimuth (3,4,5) are defined in Figure 5 and shown in practice in Figure 6. We included these variations because we wanted to ensure our dataset could train our algorithms to be invariant to changes in rotation, distance, and tensions. Although the current work only focuses on a finger-mounted solution, distance is a variable because we want to test a wrist-mounted solution in the future. The scale (6) is a division of the width of the camera's view (640 pixels) by the number of centimeters captured. We included the scale because it is helpful to have a notion of how "close" the camera is to the texture. As depicted in Figure 7, the camera can pick out individual threads of the clothing. That level of detail starts to fade in images taken at 12cm away. The lighting (7) is a measure of the power supplied to the LED (0-255 times the 5V power source squared divided by 10 ohms). This measure will prove useful in future work when we automate the LED brightness. Using the data we already have tells us what lighting level will capture consistently lit images. The tension (8) of the material is important because of the variability seen in normal conditions such as hanging in a closet or laying in a drawer. Finally, the notes (10) are simply annotations for each image and the colors (11) are the manually labelled colors of each piece of clothing.

As depicted in Figure 6, the variables and parameters chosen result in 16 possible configurations for each article of clothing considering 2 azimuth angles, 2 distances, 2 inclination angles, and 2 tensions. Notice the difference between the detail of fabric in Figure 7 and the images of Figure 8. This is one example of why the HCTD is a novel dataset with respect to clothing texture. For reference, the HCTD includes an image similar to Figure 7 for each article of clothing.

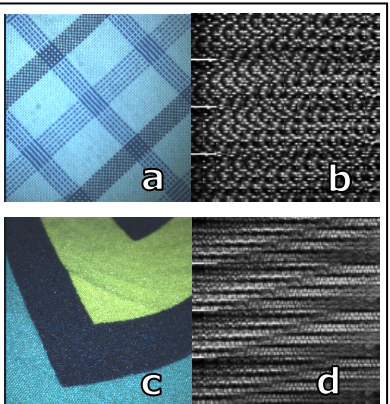

**Figure 9:** (a,b) are checkered_00038 and zigzagged_00367 images from the HCTD. (b,d) are the first of ten output matrices of the DenseSIFT algorithm, illustrated as an intensity map to demonstrate the differences between the two images.

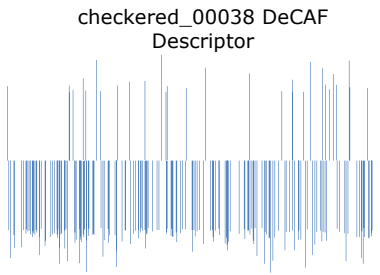

**Figure 10:** The 4096-D DeCAF feature vector which is output from the pre-trained AlexNet Caffe model after employing feature extraction on the HCTD image: checkered_00038.

As depicted in Figure 8, we have 9 texture categories which were derived from previous work in categorizing texture [1,10]. Using the previous work and eliminating all texture words such as bubbly, honeycombed, etc. (textures that do not describe clothing). We enumerated our texture categories. We also combined texture categories such as striped, banded, and lined into a single category called striped. This process was done for knitted + woven and lacelike + gauzy + frilly. Furthermore, we added denim and none to compensate for common clothing categories that did not exist in previous texture categorization work.

**Table 1:** Outlines the number of images captured for each texture, each texture's corresponding label ID, the texture type, and the number of articles of clothing for that texture.

| LabelID | Texture type | # images | # articles |
|---|---|---|---|
| 0 | Checkered | 88 | 5 |
| 1 | Denim | 40 | 3 |
| 2 | Floral | 88 | 4 |
| 3 | Knitted | 32 | 2 |
| 4 | Lacelike | 48 | 2 |
| 5 | None | 48 | 3 |
| 6 | Polka-dotted | 48 | 3 |
| 7 | Striped | 64 | 4 |
| 8 | Zigzagged | 64 | 3 |
| | **TOTAL** | **520** | **29** |

The dataset has 520 images across the 9 texture types and 29 distinct articles of clothing.

The HCTD is similar in function to the DTD [1] and the CCNY Clothing Dataset [1,6], however the HCTD is novel because it gathers close-up images of the most common clothing texture types under varied and realistic conditions. Our dataset should prove useful for any research involving clothing texture recognition. Especially considering most local clothing features extend to the entire piece of clothing. In other words, most clothes can be identified by only a few centimeters of the article.

### Texture Classification Pipeline

To classify textures we used three image features: DeCAF, Dense SIFT, and Improved Fisher Vectors. DeCAF descriptors are generated using a Deep Convolutional Neural Network (DCNN)[1]. Generally, neural networks are "trained" by feeding them hundreds of thousands of images. Furthermore, training a neural net on these large datasets, such as ImageNet[2], requires incredible computing power. In the case of low resources or small datasets, we can utilize a technique called feature extraction which uses a pre-trained neural net and the output as feature vectors.

Before elaborating on the pipeline, we'll explain how each figure represents its respective algorithm at a high-level. First an image's DeCAF vector is computed which is represented in Figures 10 and 11. Then the image is sent through DSIFT which is represented by Figure 9 (b,d). Finally, an image's DSIFT matrix is pooled into an IFV which is represented by Figures 12 and 13. Figures 10-13 have a single line for each value in the vector with x-axis being the column index in the row vector and the y-axis being the value at that index. Each value in the row vector can be positive or negative (the origin is the well-defined horizontal line in each of the Figures) and represents an attribute of an image. Notice the differences between checkered and zigzagged, using the uniqueness of each vector the SVM can "identify patterns" and use the trends to classify images.

---

[1] There are a few implementations out there, we used the Caffe Deep Learning Framework.

[2] The ImageNet is a dataset consisting of >1million images.

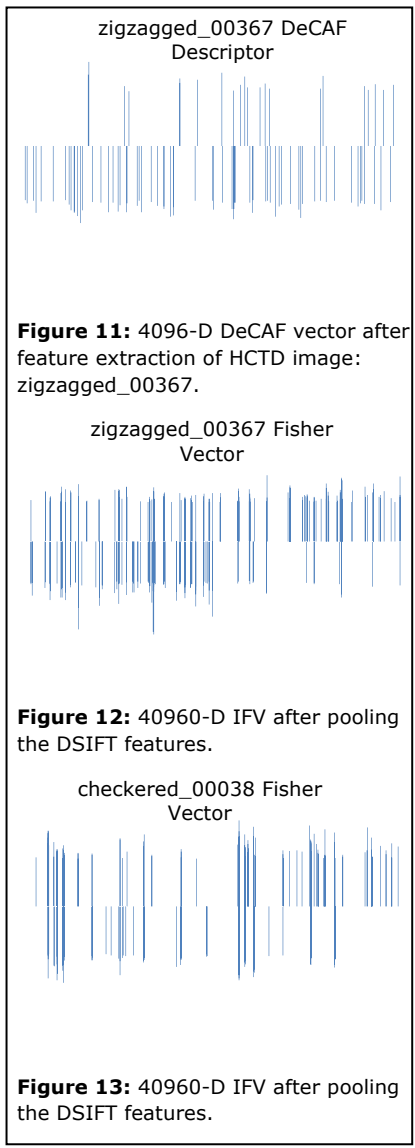

**Figure 11:** 4096-D DeCAF vector after feature extraction of HCTD image: zigzagged_00367.

**Figure 12:** 40960-D IFV after pooling the DSIFT features.

**Figure 13:** 40960-D IFV after pooling the DSIFT features.

Now, let's move onto neural networks and DeCAF. Essentially, neural nets can be imagined as a stack of layers. Each layer has inputs and outputs. Different input data will traverse the neural net along different paths. For instance, a checkered image will traverse a different path than a striped image because it "activates" different neurons. To use feature extraction, we employed the pre-trained AlexNet Caffe model and cutoff the last fully-connected layer. After the abridged AlexNet was in place, we computed the DeCAF vectors by running our images through the net and collecting the output as depicted in Figures 10 and 11. Even though AlexNet was trained on unrelated images, the output vectors are useful because the first few layers of the net identify general patterns of an image such as edges and blobs while the latter layers are more specific. By chopping the bottom layer, we compute general information in a small amount of space (4096 dimensional vectors).

The second set of our image features is captured using the Dense Scale Invariant Feature Transform (DSIFT) algorithm[3] which represents local features of an image. Figure 9 (b,d) depicts SIFT's output as an intensity map. Each column is a SIFT descriptor of size 128, and the rows are of size 196. That said, Figure 9 only shows a single iteration of the algorithm. To achieve scale invariance, the algorithm resizes the image several times and extracts descriptors on each iteration. The parameters we chose[4] result in a 128 by 392,584 element matrix. We used 10 scales ranging from 0.125 to 3 times the original image's size. More experiments are required to determine if we can reduce the number of scales while maintaining high accuracy, thereby increasing performance.

The final set of image features is the Improved Fisher Vector (IFV) which is an image representation obtained by pooling local image features such as the DSIFT descriptors as depicted in Figures 12 and 13. The IFV transforms the DSIFT matrix into a 40960-D vector which is concatenated onto the 4096-D DeCAF vector and finally fed into an SVM. We used the Dual Stochastic Coordinate Ascent (SDCA) SVM strategy.

### Evaluation of Current Work

To evaluate our classification approach we trained the SDCA SVM on the normalized IFV + DeCAF feature vectors. Then we computed the average accuracy across 40-Fold random subsampling. We performed this 40-Fold sampling at 5% increments of the HCTD used for the training set in order to find a function of accuracy versus percentage of HCTD used for training.

---

[3] We used the VLFEAT implementation of fast Dense SIFT.

[4] See source code for complete implementation and choice of parameters.

For example, at 20% HCTD used, we randomly selected 20% of the images from each texture to train the SVM. We performed 40 of these random selections and averaged the accuracy across each run. Then we incremented the amount of the HCTD used by 5% and performed 40 more randomizations. We continued this process from 20% to 80% training.

In Figure 14, notice, we have three sets of bar graphs which represent the three different feature vector permutations: Solely DeCAF vectors, solely IFV vectors, and then the combined approach IFV+DeCAF. Plotting all three lets us analyze the contribution of each strategy individually.

Now, notice how low the accuracy is for solely DeCAF vectors. These results were unexpected considering the success the DTD paper had with other datasets [1]. For example, the DTD found >90% accuracy on the UMD HR dataset using only DeCAF. This was surprising, but upon inspection of the UMD HR dataset, we see, there are 40 images per class of the same exact scene. In comparison, the HCTD has 8 images per article of clothing and has several articles per class. That said, one explanation of our poor DeCAF accuracy is simply: the HCTD is too small and there needs to be about 40 images per article of clothing to make accurate classifications. This should be relatively easy to capture as we move towards video classification because we can shoot at ~20 frames per second and gather 40 images across 2 seconds. Another detail to mention: DeCAF vectors are only 4096-D which means they carry relatively little information with respect to the IFV vectors. This means, since DeCAF vectors are smaller, there must be more images to compensate for the lack of information.

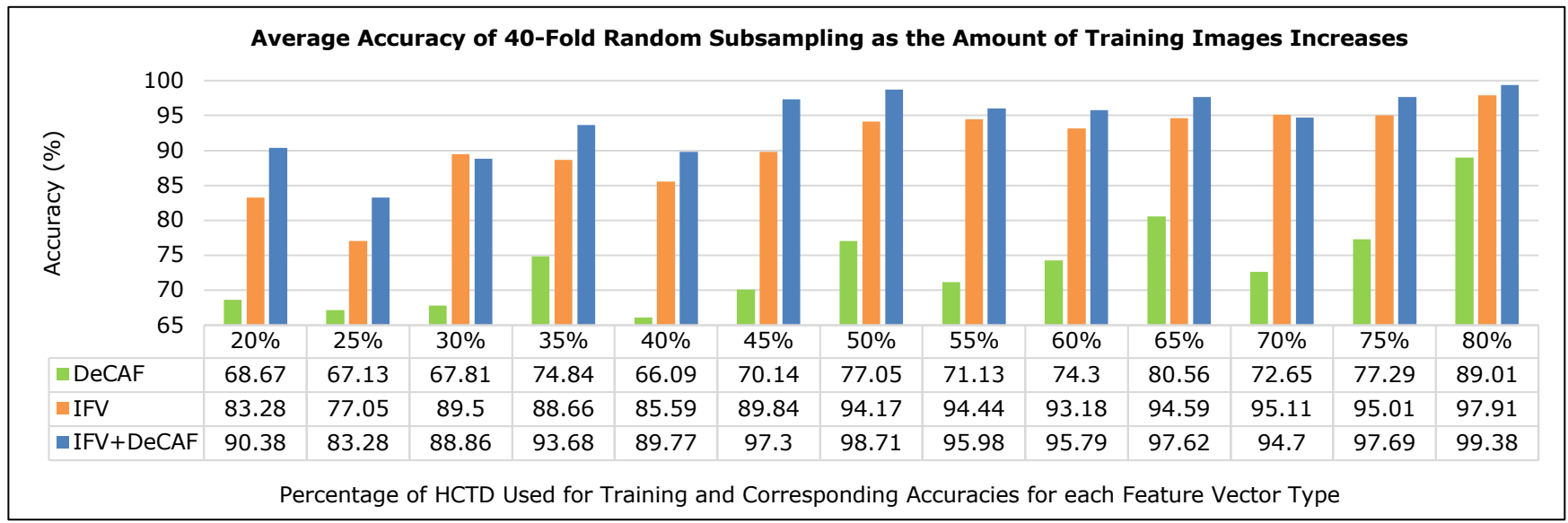

**Average Accuracy of 40-Fold Random Subsampling as the Amount of Training Images Increases**

| | 20% | 25% | 30% | 35% | 40% | 45% | 50% | 55% | 60% | 65% | 70% | 75% | 80% |
|---|---|---|---|---|---|---|---|---|---|---|---|---|---|
| DeCAF | 68.67 | 67.13 | 67.81 | 74.84 | 66.09 | 70.14 | 77.05 | 71.13 | 74.3 | 80.56 | 72.65 | 77.29 | 89.01 |
| IFV | 83.28 | 77.05 | 89.5 | 88.66 | 85.59 | 89.84 | 94.17 | 94.44 | 93.18 | 94.59 | 95.11 | 95.01 | 97.91 |
| IFV+DeCAF | 90.38 | 83.28 | 88.86 | 93.68 | 89.77 | 97.3 | 98.71 | 95.98 | 95.79 | 97.62 | 94.7 | 97.69 | 99.38 |

Percentage of HCTD Used for Training and Corresponding Accuracies for each Feature Vector Type

**Figure 14:** Depicts experimental results when using our classifier on the HCTD. There are three sets of bars here: Green, orange, and blue. Each bar represents the average accuracy of our classifier across 40 trials at each 5% interval of the HCTD used as the training set. Green represents using only the DeCAF feature vectors as input for the SVM. Orange bars represent the IFV, and blue represents IFV+DeCAF.

Now let's analyze Figure 14 in more detail. First, the difference between each 5% increment is 2-4 images per texture class. Second, there is an upward trend: as the percentage of HCTD used for training increases - the accuracy increases. Finally, notice the dips at 25%, 40%, 55%, and 70%. These dips are equally spaced which could mean there is a deeper meaning to the dips, but for now, we'll write it off as a poorly behaved function. In fact, statistically, the 25% dip is an outlier and the 30% is 0.4% accuracy from being an outlier.

Perhaps the most important theme to extract from this is the following: there is not enough data per class and per piece of clothing to provide enough information for consistent classifications. For example, notice the unpredictability of the IFV (orange) until ~40% of the dataset. After which point, the IFV accuracies stabilize and follow a predictable upward trend. The DeCAF, however, remains unpredictable throughout all 13 increments of HCTD used. That makes sense because the DeCAF vectors contain a tenth the amount of information as the IFV vectors. By that logic, we will need approximately 10 times the amount of images for the DeCAF vectors to behave predictably. These results demonstrate how much data is required to create accurate classifications using a neural net.

## Discussion and Conclusion

The HandSight Color-Texture Dataset was successful in providing an image set within the problem domain (close-up, consistent lighting, and varied tensions). However, the dataset is currently too small. To build a more general classifier we need at least a few hundred pieces of clothing and greater than 16 variations in perspective. Moving forward, we plan to build an automatic solution that employs a robotic arm which changes the camera orientation about a fixed position. We will also use local thrift stores to gather a broader range of clothing. Applying both of these solutions in some form are required to build a dataset that can train a sufficiently robust classifier (human-level accuracy across unknown sets of clothing). That said, we need to evaluate the current classifier on unknown clothes to determine how close we are to a realistic solution. Finally, the DTD classifier is trained on more than 5000 images, we should evaluate the HCTD with the DTD model. Perhaps the DTD model is able to classify the HCTD with high accuracy eliminating the need for more elaborate image collection strategies.

Aside from the HCTD, this work evaluated an image classification pipeline consisting of IFV and DeCAF feature vectors which classified clothing with a >95% accuracy. Moving forward we will fine-tune the Caffe model with the HCTD which we expect to improve accuracy with almost no performance cost. A large reason we didn't use the Caffe framework exclusively is a lack of input data. Training a neural net requires extensive datasets which wasn't feasible in this work. Perhaps, once we move forward with the automatic data collection schema we will have a dataset large enough to use Caffe alone. Caffe requires much less computation than DSIFT/IFV which means if Caffe produces high accuracy on a well-trained model, exclusive Caffe use could lead to real-time performance. To illustrate that point further, it takes ~180ms per image for DeCAF and ~3.8s per image for DSIFT/IFV.

Achieving high accuracy classification is certainly a milestone, however, we need to research more precise user feedback mechanisms as well. For example, Figure 15 depicts a green, pink, and white, floral skirt, however, a simple color-texture description does not convey enough meaning for a blind user to fully understand the article of clothing.

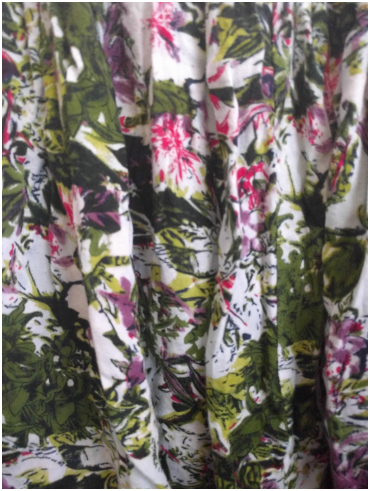

**Figure 15:** An example of a piece of clothing that requires more description than mere color and texture to present an accurate description to a user.

Overall, this work uses a variant of the texture classification used in the DTD paper and reaches similar accuracies of >95%. Our texture recognition solution is approaching that of human accuracy, however the speed of which needs improvement. Our pipeline runs on the order of seconds per image with an Intel i7 processor and 8GB of RAM. Further research is required to build a parallelized solution that can run on embedded GPUs.